\DocumentMetadata{}
\documentclass[acmsmall]{acmart}
\AtBeginDocument{%
  }

\setcopyright{acmcopyright}
\copyrightyear{2026}
\acmYear{2026}
\acmDOI{XXXXXXX.XXXXXXX}

\usepackage{hyperref}
\usepackage{url}
\usepackage{multirow}
\usepackage{enumitem}
\usepackage{float}
\restylefloat{table}
\usepackage{wrapfig}
\usepackage{tikz}
\usepackage{todonotes}
\usepackage{tabularx}
\usepackage{graphicx}
\usepackage{wrapfig}

\begin{document}

% old title:
%\title[NLP to Enhance Deliberation in Political Online Discussions: A Survey]{Natural Language Processing to Enhance Deliberation in Political Online Discussions: A Survey}

\title[ML for Enhancing Deliberation in Online Political Discussions and Participatory Processes: A Survey]{Machine Learning for Enhancing Deliberation in Online Political Discussions and Participatory Processes: A Survey}

\author{Maike Behrendt}
\email{maike.behrendt@uni-duesseldorf.de}
\orcid{0009-0000-4227-7743}
\affiliation{%
	\institution{Heinrich Heine University D{\"u}sseldorf}
	\streetaddress{Universit{\"a}tsstraße 1}
	\city{D{\"u}sseldorf}
	\country{Germany}
	\postcode{40225}
}

\author{Stefan Sylvius Wagner}
\affiliation{%
	\institution{Heinrich Heine University D{\"u}sseldorf}
	\streetaddress{Universit{\"a}tsstraße 1}
	\city{D{\"u}sseldorf}
	\country{Germany}}
\email{stefan.wagner@uni-duesseldorf.de}

\author{Carina Weinmann}
\affiliation{%
	\institution{Heinrich Heine University D{\"u}sseldorf}
	\streetaddress{Universit{\"a}tsstraße 1}
	\city{D{\"u}sseldorf}
	\country{Germany}}
\email{c.weinmann@uni-duesseldorf.de}

\author{Marike Bormann}
\affiliation{%
	\institution{Heinrich Heine University D{\"u}sseldorf}
	\streetaddress{Universit{\"a}tsstraße 1}
	\city{D{\"u}sseldorf}
	\country{Germany}}
\email{marike.bormann@uni-duesseldorf.de}

\author{Mira Warne}
\affiliation{%
	\institution{Heinrich Heine University D{\"u}sseldorf}
	\streetaddress{Universit{\"a}tsstraße 1}
	\city{D{\"u}sseldorf}
	\country{Germany}}
\email{mira.warne@uni-duesseldorf.de}

\author{Stefan Harmeling}
\affiliation{%
	\institution{Technical University Dortmund}
	\streetaddress{August-Schmidt-Straße 4}
	\city{Dortmund}
	\country{Germany}
	\postcode{44227}}
\email{stefan.harmeling@tu-dortmund.de}

\renewcommand{\shortauthors}{Behrendt et al.}

\begin{abstract}
	Political online participation in the form of discussing political issues and
	exchanging opinions among citizens is gaining importance with more and more
	formats being held digitally. To come to a decision, a thorough discussion and 
	consideration of opinions and a civil exchange of arguments, which is defined
	as the act of \emph{deliberation}, is desirable. The quality of discussions
	and participation processes in terms of their deliberativeness highly depends
	on the design of platforms and processes. To facilitate online communication
	for both participants and initiators, machine learning methods offer a lot of
	potential. In this work we want to showcase which issues occur in political
	online discussions and how machine learning can be used to counteract these
	issues and enhance deliberation. We conduct a literature review to (i) identify tasks that could potentially be solved by artificial intelligence (AI) algorithms to enhance individual aspects of deliberation in political online discussions, (ii) provide an overview on existing tools and platforms that are equipped with AI support and (iii) assess how well AI support currently works and where challenges remain.
\end{abstract}




\received{20 September 2023}
\received[revised]{04 January 2026}
\received[accepted]{}

\maketitle

\section{Introduction}

Over the last two decades, the amount of research has grown continuously, particularly with the
ongoing emergence of new platforms that allow for exchange online. Besides simple
exchange of opinions, online participation platforms are also effective
instruments to make use of collective intelligence to find responses to severe
challenges like pandemics~\citep{mouter2021public} or climate change~\citep{5928663}.

Online political participation comes in many different forms. While some
definitions include both passive (e.g.,
reading news articles) and active (e.g., voting) political behavior, this
review focuses exclusively on active engagement through written opinion exchange, e.g., in online citizen councils, referendums and
discussions on political topics~\citep{ruess2021online}. This also includes informal exchanges on social media or news websites, which have been shown to stimulate citizens' political engagement~\citep{conroy2012facebook, durotoye2025online}.
From a deliberative perspective, participation in such discussions can be seen as an essential, if not central aspect of citizens' democratic involvement 
\citep{doi:10.1080/17448689.2013.871911,
	doi:10.1146/annurev.polisci.7.121003.091630,
	https://doi.org/10.1111/j.1475-6765.2006.00636.x}.

% importance of social media debates for political participation

The theoretical foundation for this perspective is deliberative democracy, which has become one of the most influential conceptions of democracy in
political theory and research~\citep{doi:10.1177/0192512120941882}. It represents a
normative approach, under which basically a multitude of deliberative theories
can be subsumed, each with a different
emphasis~\citep{bohman1997deliberative, cohen1989good,
	dryzek2002deliberative, fishkin2008debating, gutmann2004deliberative}.
However, they agree that democracy should consist of a discursive exchange among
equals with the participation of all members of society. This exchange should be
characterized by rational arguments as well as the willingness to
respectfully acknowledge and understand the viewpoint and arguments of the other
side, to reconsider and, if necessary, to change one's own viewpoint. The goal
of this exchange is to jointly find solutions to social
problems~\citep{chambers2003deliberative, cohen1989good,
	https://doi.org/10.1111/j.1467-8675.1994.tb00001.x}.

Deliberation can be characterized by the following four core dimensions~\citep{friess2015systematic}:
\begin{itemize}%[nosep]
	\item \emph{Rationality}: argumentative exchange of opinions.
	\item \emph{Interactivity}: being responsive and listening to each
	other. 
	\item \emph{Equality}: same opportunities for anyone.
	\item \emph{Civility}: respectful interaction between participants.
\end{itemize}

The degree to which online discussions fulfill these deliberative criteria depends strongly on both platform design~\citep{esau2017design} and the presence of effective moderation. Human moderators can substantially improve discussion quality~\citep{doi:10.1080/10584609.2020.1830322,10.1145/3555095}, yet moderation is a resource-intensive task that is difficult to scale across large online platforms~\citep{10.1145/3579530}.
% hier übergang zu Moderator*innen
Recent advances in AI, and particular in machine learning (ML), and natural language processing (NLP), offer promising opportunities to support and enhance online deliberative processes in online discussions~\citep{argyle2023leveraging}. ML methods can help structure large-scale exchanges, identify patterns in argumentation or sentiment, and foster more civil and constructive dialogue. NLP, as a subfield of ML and computational linguistics, provides tools for analyzing and generating human language, thereby enabling automated facilitation and moderation of online discussions. The technical terms such as ML and NLP are defined in detail in Section \ref{subsec:term}.

\paragraph{Scope and Focus of This Review}
The overarching goal of this paper is to explore how ML and NLP can contribute to enhancing deliberation in online political discussions.
Specifically, we aim to: 
\begin{itemize}%[nosep]
	\item provide a comprehensive overview of common issues in online
	discussions and participation processes,
	\item identify and describe ML and NLP tasks that could enhance deliberative quality in	online discussions,
    \item summarize existing studies examining the effects of ML based interventions,
	\item review current ML and NLP techniques applied
	in online discussion contexts,
	\item and outline open challenges and directions for future research.
\end{itemize}

From these objectives, we derive the following research questions: 
\begin{enumerate}
    \item What are the main issues that occur in political online discussions and participatory processes?
    \item Which tasks can potentially be addressed with ML or NLP algorithms?
    \item How can ML and NLP contribute to enhancing deliberation?
    \item How well do algorithms perform at the moment and what challenges remain?
\end{enumerate}

This review concentrates on political online discussions and participation processes. We exclude applications in online education and collaborative learning platforms, where objectives and user dynamics differ substantially (for an overview, see~\citet{OUYANG2024100616}). Similarly, tools designed for group meetings or seminars, such as the discussion board system by~\citet{sasaki2021online}, fall outside our scope. Given the particular importance of transparency and accessibility for democratic participation, our review focuses on open-source or non-profit platforms that are freely available to the public.
Overall, we identified 36 tasks within five sub-categories that could be solved
by ML methods to help to enhance deliberation in online discussions. 
We found 23 software tools that have been developed or are still being developed that already incorporate NLP to enhance online discussions.
Together, these findings provide a systematic overview of the state of research and technology in this emerging field. 

Our paper is structured as follows: first, we describe the methodology employed for conducting the extensive literature review in Section \ref{sec:meth}. 
We then define the most important terms in Section \ref{subsec:term} and discuss known issues in both online participation processes as well as general online discussions in Sections~\ref{subsec:issues} and \ref{subsec:disissues}. Next, we provide an overview of the tasks that can be solved by ML methods to enhance deliberation in online discussions in Section~\ref{sec:tasks}. Subsequently, we present software tools that already make use of NLP methods in practice in Section~\ref{sec:platforms}. Finally, we outline key challenges and open problems in Section \ref{sec:challenges}.
Section~\ref{sec:conlcusion} concludes the paper with a summary of our findings and an outlook on future research directions.

\begin{wraptable}{r}{5cm}
	\centering
    \vspace{-1em}
	\begin{tabular}{|l|}
		\hline
		"AI" / "NLP" / "Machine Learning"\\
		+ "citizen councils"\\
		+ "citizen participation"\\
		+ "decision making" \\
		+ "e-participation"\\
		+ "group decision making" \\
        + "group discussions" \\
		+ "online deliberation"\\
		+ "online discussions"\\
        + "moderation" \\
        + "social media" \\
		+ "social sciences" \\
		"facilitation of online discussions"\\
        "issues in online discussions" \\
		\hline
	\end{tabular}
	\caption{Used search terms for the conducted literature review.}
	\label{tab:search_terms}
\end{wraptable}

\section{Methodology}
\label{sec:meth}
We employed an iterative literature review process, which we describe below. A comprehensive search was conducted across relevant databases in NLP, with academic search engines such as Google Scholar serving as a starting point for a systematic, forward and backward snowballing process, as outlined in \citep{10.1145/2601248.2601268}. Snowballing, also known as citation chaining, is the process of identifying relevant works and examining the cited literature, as well as the literature that has cited the relevant works, to locate additional significant papers. To ensure the timeliness of the research, the study exclusively considered publications subsequent to 2014. We did not restrict our search to a certain research field, as the topic is highly interdisciplinary. %During the search, a framework was employed to sort the found literature into categories that are closely related to the dimensions of deliberation, including \emph{inclusiveness}, \emph{rationality}, \emph{civility}, \emph{reciprocity}, and \emph{others}. 
The criteria established for potential tasks to solve in order to enhance online deliberation were rather broad. In the case of discussion and political participation platforms we excluded platforms that were not (1) open source, or non-profit, (2) supported by any kind of AI, and (3) developed for online discussions, deliberation and participation processes, without a special focus on, e.g., online education and business meetings. Additionally, the Democracy Technologies Database\footnote{\url{https://democracy-technologies.org/database/}} was utilized to identify discussion tools and platforms. In total, we identified 36 tasks, which we assigned to 5 different categories, as well as 23 software tools for online discussions or their evaluation, and 22 datasets for training and evaluating ML models on discussion data.

\paragraph{Search Terms}
Table \ref{tab:search_terms} lists search terms we used in different
combinations to gain a broad overview on the topic of ML and NLP to support
political online discussions and online participation processes.

\section{Background}
\label{sec:background}

Before discussing tasks that can be solved by NLP to support participants and initiators
of online participation processes and online discussions in general, we want to first provide the reader with a definition of AI, ML, NLP, Large Language Models and deliberation, and afterwards shed light on challenges in both contexts to better understand how ML and NLP can be deployed. 

\begin{figure}
    \includegraphics[width=0.7\textwidth]{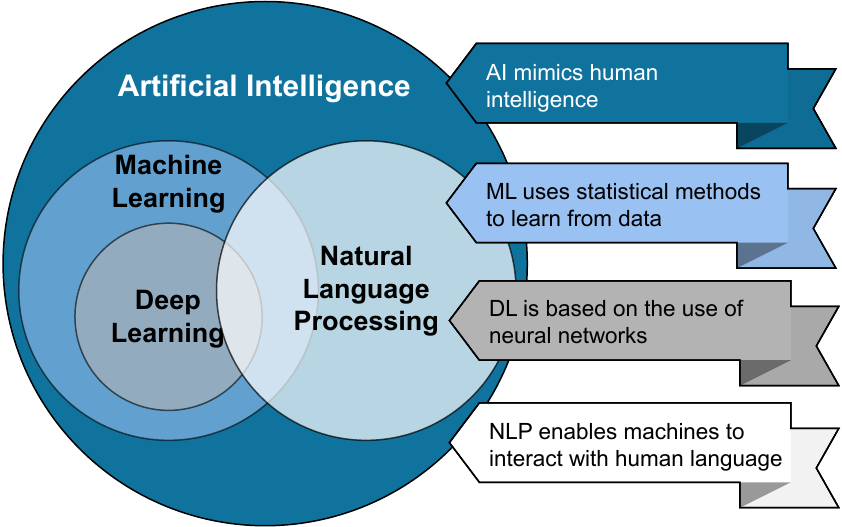}
    \caption{Classification of Artificial Intelligence related terms.}
    \label{fig:illustration}
\end{figure}

\subsection{Terminology}
\label{subsec:term}
This review is aimed at computer scientists, as well as social and political scientists, as the reviewed literature comes from a highly interdisciplinary field where technology and the theory of deliberation converge. In order to define the technical terms used in this paper, they are explained in detail below.

\paragraph{Artificial Intelligence} AI is the study and design of algorithms that enable machines to perform tasks that normally require human intelligence, such as learning, reasoning, and problem-solving.

\paragraph{Machine Learning}
ML is a branch of AI that enables computers to automatically learn from data and improve their performance without explicit programming. ML bridges computer science, mathematics, and statistics, and provides the foundation for many applications such as pattern recognition, predictive modeling, and NLP.

\paragraph{Deep Learning} DL is a branch of ML. Traditional ML methods often rely on humans to select features that are fed into algorithms such as decision trees. In contrast, deep learning is based on artificial neural networks that learn features directly from the data~\citep{janiesch2021machine}. Deep learning uses layered neural networks which are modeled on the structures of the human brain. These networks can automatically discover important patterns in large amounts of data, e.g., learning to recognize objects in images without being told which features to look at. This makes deep learning effective for tasks such as speech recognition, language understanding, and image classification.

\paragraph{Natural Language Processing}
NLP is a subfield of computer science and computational linguistics that develops methods to enable machines to process and interact with human language. The field covers a broad spectrum of tasks, including semantic text analysis, sentiment detection, morphological segmentation, machine translation, and text summarization. In the context of online discussions, text contributions are the main data, which is why NLP is relevant here. Figure \ref{fig:illustration} illustrates the relation between AI, ML, DL and NLP. 

\paragraph{Large Language Models}
Large Language Models (LLMs) are a recent development in NLP, based on DL. LLMs are deep neural networks that are trained on extensive collections of text. They learn to statistically represent linguistic structures and meanings. Consequently, LLMs can generate and interpret language in a remarkably flexible and human-like way. Because they can generate content, LLMs are considered \emph{generative AI}. Building upon advances in m ML, LLMs have become central to applications such as chatbots, text summarization, and automated writing assistance.

\paragraph{Deliberation}
Deliberation can be understood as a communicative process in which participants respectfully and thoughtfully exchange arguments with the goal of forming a collective judgment. It is commonly described along four dimensions: \emph{rationality}, which refers to the logical and evidence-based justification of positions; \emph{civility}, which encompasses respect and politeness, \emph{interactivity}, which highlights engagement and responsiveness in dialogue; and \emph{equality}, which ensures that every participant has the same opportunity to participate~\citep{bachtiger2009measuring, esau2021different, 10.1007/978-3-642-15158-3_3}.

%\citet{ORBi-ca9c331b-d6ec-426d-9db3-4a17bb7ab193} analyzed citizen participation
%processes in two cities in Belgium both using customized versions of the
%CitizenLab platform. They point out the following issues: used NLP methods
%should be transparent, else they can lead to a bias of the whole participation
%(example: keyword extraction and automatic generation of topic labels).
%Duplicate proposals can also bias the process as votes for ideas are distributed
%over similar ideas, also duplicates lead to a higher number of proposals that
%are simply too many to go through in detail.
\subsection{Issues in Online Participation Processes}
\label{subsec:issues}
We will take a closer look at issues that have been reported for online participation processes in the following. The first challenge that initiators of participation processes have to deal with 
are \emph{low rates of participation}, due to various reasons such as lack of motivation, lack of awareness about the process, or lack of online access~\citep{shortall2022reason, 10.1145/3428502.3428614}. Promoting public participation projects and ensuring their success is, however, a task that is out of scope for NLP methods, as it takes place before the actual process of participation. Another phenomenon in participation processes, in which citizens can make proposals or share ideas, is the large number of gathered posts. Too many contributions cause what is referred to as \emph{information overload}, which is denoted as one of the key challenges for digital mass participation by \citet{10.1145/3452118}. One effect of information overload is the \emph{redundancy of contributions} which makes it difficult for participants to find proposals of interest to support or comment~\citep{10.1145/3428502.3428614, ORBi-ca9c331b-d6ec-426d-9db3-4a17bb7ab193}.
Being confronted with a multitude of postings on a platform makes it impossible for participants to gain an overview of ideas that have been added and leads to \emph{a lack of interaction} as they solely share their own ideas without noticing others. As the number of postings grows, it also becomes difficult for local authorities to \emph{evaluate} the process afterwards~\citep{ORBi-ca9c331b-d6ec-426d-9db3-4a17bb7ab193}. 

\citet{SIMONOFSKI2021106901} analyze requirements of both participants and public servants for a participation process
in Belgium and elaborate guidelines for the development and implementation of
participation platforms. One guideline includes the consideration of NLP support
to cluster and summarize submitted ideas. \citet{dian2019ai} examine how the
application of NLP in online citizen participation could be standardized. They point out the importance of high-quality data for training models in order to make adequate predictions. They also state that human based analysis is still preferable for
some tasks as humans perform much better on the explored NLP tasks, while
additionally the decisions of a ML method are often not explainable.

\citet{Tsai2024Generative} present a framework for what to consider and what guiding principles to follow when integrating generative AI into pro-democratic citizen participation platforms. They see great potential in augmenting processes with AI to allow for more informed processes in which participants have equal opportunities to participate, provided that the user's agency and autonomy are not compromised. However, they also point out fundamental risks, such as potential over-censorship by AI agents, the misuse of AI to generate deepfakes, and over-reliance on AI.

\subsection{Issues in Other Forms of Online Discussions}
\label{subsec:disissues}
Several reports and studies have examined issues and drawbacks in online
discussions on social media platforms in general. Since these issues can likely be transferred to political contexts, we will summarize them in the following.

The reported issues include \emph{disorganized content}, \emph{redundancy in posts}, \emph{many contributions of low quality}, and \emph{polarization}~\citep{klein2011harvest}.
Other issues are \emph{formation of small groups} that share the same opinion, which is connected to the emergence of filter bubbles and echo chambers, \emph{poor argumentation}, which leads to
\emph{biases} and a \emph{lack of references} to the contributions of other participants~\citep{klein2015critical, 10.5555/3535850.3536112, gao2013designing}, and incivility toward other users~\citep{anderson2014nasty}.
Many of the mentioned problems in online discussions are related to one or multiple dimensions of deliberation.
Human facilitators are responsible for managing discussions and preventing the aforementioned issues. However, the growing number of participants and accompanying increase in posts make it difficult to maintain an overview~\citep{ito2014collagree}. NLP methods could support and simplify the work of human facilitators and initiators of political online discussions. Selecting the appropriate tools and facilitation mechanisms is a crucial part of digital online participation
in order to conduct a successful process and rule out possible negative side effects~\citep{simon2017digital}. %In the following, we give an overview of tasks that can be solved by ML methods to support facilitators, participants, and initiators of online discussions.

% OTHER POTENTIAL TASKS:
% CONTENT RECOMMENDATION: e.g. topics, comments, users
% CONSENSUS DETECTION? SUMMARIZATION OR ARGUMENT MINING??

\begin{figure*}
	%\footnotesize
	\centering
	\begin{tikzpicture}[>=stealth, 
		category/.style={shape=rectangle, align=center,text width=3.1cm, minimum height=1.01cm, node distance=0.1cm and 0.25cm}, %font=\bfseries
		method/.style={shape=rectangle,draw, align=center,text width=2.75cm, fill=gray!30, minimum height=0.75cm, node distance=0.1cm and 2cm},
		submethod/.style={shape=rectangle,draw, align=center,text width=2.25cm, fill=white!10, minimum height=0.75cm, node distance=0.1cm and 0.5cm}]
		
		% create the nodes
        %GENERATING INPUT
		\node (gi) [category]{\hyperref[sec:gen]{Section \ref{sec:gen}}\\ \textbf{Generating Input}};
		\node (at) [method, below=of gi]{\hyperref[par:trans]{Automatic Translation}};
		\node (amo) [method, below=of at]{\hyperref[par:mod]{Automatic Moderation}};
		\node (mop) [submethod, below=of amo, xshift=0.25cm]{Moderation Intervention Prediction};
		\node (rui) [submethod, below=of mop]{Rephrasing User Input};
		\node (sob) [submethod, below=of rui]{Social Bot Detection};
        \node (ssd) [submethod, below=of sob]{Social Spam Detection};
		\node (fc) [method, below=of ssd, xshift=-0.25cm]{\hyperref[par:fact]{Fact-Checking}};
		\node (rud) [submethod, below=of fc, xshift=0.25cm]{Rumor Detection};
		\node (ere) [submethod, below=of rud]{Evidence Retrieval};
		\node (vpr) [submethod, below=of ere]{Verdict Prediction};

        % STRUCTURING THE DISCUSSION
		\node (sdi) [category,right=of gi]{\hyperref[sec:struc]{Section \ref{sec:struc}} \\ \textbf{Structuring the Discussion}};
		\node (ami) [method, below=of sdi]{\hyperref[par:argu]{Argument Mining}};
		\node (iaa) [submethod, below=of ami, xshift=0.25cm]{Inter-Argument Agreement};
		\node (ped) [submethod, below=of iaa]{Personal Experience Detection};
		\node (kpa) [submethod, below=of ped]{Key Point Analysis};
		\node (dr) [method, below=of kpa, xshift=-0.25cm]{\hyperref[par:dup]{Duplicate Recognition}};
		\node (ds) [method, below=of dr]{\hyperref[par:summ]{Dialogue Summarization}};
		\node (tsi) [submethod, below=of ds, xshift=0.25cm]{Text Simplification};
        \node (cod) [submethod, below=of tsi]{Consensus Detection};
		\node (ste) [method, below=of cod, xshift=-0.25cm]{\hyperref[par:struc]{Structure Extraction}};
		\node (tm) [method, below=of ste]{\hyperref[par:top]{Topic Modeling}};
        \node (crm) [submethod, below=of tm, xshift=0.25cm]{Content Recommendation};
		
		\node (as) [category,right =of sdi]{\hyperref[sec:sent]{Section \ref{sec:sent}}\\ \textbf{Analysis of Sentiment}};
		\node (sean) [method,below =of as]{\hyperref[sec:sent]{Sentiment Analysis}};
		\node (ipl) [submethod, below=of sean,xshift=0.25cm] {Identifying Opinion Leaders};
		\node (ppd) [submethod, below=of ipl] {Politeness Prediction};
		\node (hsd) [method, below=of ppd,xshift=-0.25cm]{\hyperref[par:hate]{Hate Speech Detection}};
		\node (sad) [submethod, below=of hsd,xshift=0.25cm] {Sarcasm Detection};
		\node (rad) [submethod, below=of sad] {Racism Detection};
		\node (sed) [submethod, below=of rad] {Sexism Detection};
		\node (sde) [method, below=of sed,xshift=-0.25cm]{\hyperref[par:stance]{Stance Detection}};
		\node (pvi) [submethod, below=of sde,xshift=0.25cm]{Political Viewpoint Identification};
		
		%measuring quality
		\node (mdq) [category,right =of as]{\hyperref[sec:qual]{Section \ref{sec:qual}}\\ \textbf{Deliberative Quality}};
		\node (dq) [method, below=of mdq]{\hyperref[par:qual]{Deliberative Quality Prediction}};
		\node (aqp) [method, below=of dq]{\hyperref[par:argqual]{Argument Quality Prediction}};
		
		%evaluation
		\node (ev) [category,below=of aqp, yshift=-0.1cm]{\hyperref[sec:evaluation]{Section \ref{sec:evaluation}} \\\textbf{Evaluation}};
		\node (eval) [method, below=of ev]{\hyperref[sec:evaluation]{Evaluation}};
		\node (amin) [submethod, below=of eval,xshift=0.25cm]{Argument Mining};
		\node (san) [submethod, below=of amin]{Sentiment Analysis};
        \node (sum) [submethod, below=of san]{Summarization};
		%	
		%%	 connect the nodes
        % AUTOMATIC MODERATION
		\draw[->] ([xshift=-1.25cm]amo.south) |- (mop.west) ; 
		\draw[->] ([xshift=-1.25cm]amo.south) |- (rui.west) ; 
		\draw[->] ([xshift=-1.25cm]amo.south) |- (sob.west) ; 
        \draw[->] ([xshift=-1.25cm]amo.south) |- (ssd.west) ; 
        	
		% FACT CHECKING
		\draw[->] ([xshift=-1.25cm]fc.south) |- (rud.west) ; 
		\draw[->] ([xshift=-1.25cm]fc.south) |- (ere.west) ; 
		\draw[->] ([xshift=-1.25cm]fc.south) |- (vpr.west) ; 

        % ARGUMENT MINING
		\draw[->] ([xshift=-1.25cm]ami.south) |- (iaa.west) ; 
		\draw[->] ([xshift=-1.25cm]ami.south) |- (ped.west) ; 
		\draw[->] ([xshift=-1.25cm]ami.south) |- (kpa.west) ; 

        % DIALOG SUMMARIZATION
		\draw[->] ([xshift=-1.25cm]ds.south) |- (tsi.west) ; 
        \draw[->] ([xshift=-1.25cm]ds.south) |- (cod.west) ; 

        % TOPIC MODELING
		\draw[->] ([xshift=-1.25cm]tm.south) |- (crm.west) ;         
        
        % SENTIMENT ANALYSIS
		\draw[->] ([xshift=-1.25cm]sean.south) |- (ipl.west) ; 
        \draw[->] ([xshift=-1.25cm]sean.south) |- (ppd.west) ; 

        % HATE SPEECH DETECTION
		\draw[->] ([xshift=-1.25cm]hsd.south) |- (sad.west) ; 
		\draw[->] ([xshift=-1.25cm]hsd.south) |- (rad.west) ; 
		\draw[->] ([xshift=-1.25cm]hsd.south) |- (sed.west) ; 

        % STANCE DETECTION
		\draw[->] ([xshift=-1.25cm]sde.south) |- (pvi.west) ; 

        % EVALUATION
		\draw[->] ([xshift=-1.25cm]eval.south) |- (amin.west) ; 
		\draw[->] ([xshift=-1.25cm]eval.south) |- (san.west) ; 
        \draw[->] ([xshift=-1.25cm]eval.south) |- (sum.west) ;

	\end{tikzpicture}
	\caption{Identified NLP tasks that could potentially support and improve online discussions.}
	\label{fig:identified_tasks}
\end{figure*}

\section{Tasks to Solve in Online Discussions}
\label{sec:tasks}

During the review of literature on the topic of NLP and ML to support online discussions, we identified a list of tasks with great potential for facilitating and supporting online discussions. In the following, we explain each task and list important literature. Please note that, since a large body of literature exists for each task, we only discuss significant work or refer to recent overviews summarizing the literature on specific topics. If available, we also provide evidence showing effectiveness of these tasks when applied to online discussions. However, it is important to note that the usefulness of an AI tool applied in an online discussion depends on multiple factors. While the performance of the tool is important, the design of the platform is equally relevant \citep{esau2017design}.

Many of the tasks discussed in this section were initially considered classification tasks that relied on annotated data to train supervised classification models. With the rise and improvement of large generative language models such as GPT-5\footnote{\url{https://openai.com/index/introducing-gpt-5/}}, a new paradigm has established and much research evaluates the capacities of these generative models to solve NLP tasks to enhance deliberation~\citep{wachsmuth-etal-2024-argument} and to generate and augment data to train models~\citep{wagner2025the}.

We categorize the identified tasks (see Figure \ref{fig:identified_tasks} for an overview) into the following categories: (i) tasks that generate input, (ii) tasks to structure the discussion, (iii) tasks to analyze the
sentiment, (iv) tasks that measure discourse quality and (v) tasks that help evaluate the discussion. Some tasks appear multiple times, as, e.g., sentiment analysis and argument mining can be applied during a discussion, as well as during the evaluation phase.

\subsection{Generating Input}
\label{sec:gen}

Probably one of the most direct ways NLP can be used in discussions is to let a
model generate input for the discussion. In the following, we take a look at
NLP tasks that enhance the discussion by generating input in some way. This can
range from simple automatic translations to more complex tasks like automated
moderation.

\paragraph{Automatic Translation.}
\label{par:trans}
Automatic translation of texts holds great potential for online deliberation, as
it allows for more inclusiveness and thereby \emph{equality} of participants, by
breaking down language barriers. If high-quality translations can be provided in real time
for any language, more people with different cultural backgrounds can engage in decisions and share their opinion.
With the emergence of
the Transformer model by \citet{NIPS2017_3f5ee243}, automatically generated translations have become fairly reliable and are currently mostly researched for low resource languages.
\citet{anastasiou-2022-enrich4all} for example proposes a BERT-based~\citep{devlin_etal_2019_bert} language model for Luxembourgish that will be applied in public administration as a chatbot for automatic translation and question answering. \citet{stahlberg2020neural} provides a recent overview on methods and applications for automatic translation.
% Research is focusing on multilingual models \citep{} and translation for low
% resource languages \citep{].
	
	\paragraph{Automatic Moderation.}
    \label{par:mod}
	Moderation of online discussions is a crucial task that often determines the
	quality of held discussions~\citep{doi:10.1080/10584609.2020.1830322}. It is a time and resource 
	consuming task, that could be facilitated or even replaced by artificial moderators. 
	As an initial approach to support human moderators, \citet{falk-etal-2021-predicting} train models on the task of moderation intervention prediction on
	the Regulation Room Corpus~\citep{park2012facilitative}. 
	They aim to classify whether a comment needs intervention, which can then be suggested to a human moderator.
    In their study, \citet{10.1145/3543507.3583275} examine the effects of fully automated content moderation. They apply a system that automatically hides or deletes comments that break community guidelines. They find that automatic deletion of comments has a positive effect on the posting behavior of users. Only hiding content did not obtain the same effect. Similar effects have been found when intervening even before posting harmful content~\citep{katsaros2022reconsidering}. \citet{argyle2023leveraging} use a GPT-3-based~\citep{brown2020language} chat assistant in one-on-one discussions about gun control that suggests rephrased messages in real time. Their study shows that the quality of conversations and the feeling of being understood by their counterpart could be improved through the application of the ML assistant. Checking user comments for their tone and suggesting reformulations can increase \emph{civility} of discussions, creating a better atmosphere in which more people tend to express their opinions. 
    \citet{klein2025can} suggests to let LLMs generate more precise user contributions in cases where they lack context or use complex terms that not everyone understands which contributes to the \emph{equality} of each participant within the discussion and could also improve \emph{interactivity} between users.

	There are many other examples of recent developed chatbots, e.g., to structure the conversation~\citep{10.1145/3449161}, to identify harmful language~\citep{Clever2022, electronics13030544} and spam \citep{electronics13030544}, or to interact with participants by posting messages~\citep{ito2022agent}.
	
	The technology behind supportive automated moderators is the same as that used by social bots. The difference between the two is their purpose. Social bots appear as human profiles on social media, however, they are used to spam, spread misinformation, and manipulate users on social media platforms, such as Twitter~\citep{doi:10.1177/0894439320908190}.
	Therefore, an important task for participation processes and social media
	discussions is detecting social bots~\citep{9505695} and social spam \citep{RAO2021115742}.
    
    Recent studies show that automated content moderation has a great potential in reducing harmful content on online discussion platforms, which enhances deliberative values, especially \emph{civility}. AI-assisted moderation was found to motivate participants to write more constructive comments and follow the given rules on online platforms \citep{yu2024systematic}, enhancing both \emph{interactivity} and \emph{rationality}. Nevertheless, the use of AI models should be transparent, and users should be informed why their comment was deleted. LLMs can perform this reasoning process~\citep{10.1145/3613905.3650828}, but human moderators currently make the most comprehensible decisions and can therefore be supported by AI, without being completely replaced. For a recent systematic literature overview on the effects of automatic content moderation see the work of \citet{yu2024systematic}.
	
	\paragraph{Fact-Checking.}
    \label{par:fact}
	Fact-checking is a crucial task in online discussions, as the spread of misinformation online
	is a severe problem~\citep{rocha2021impact}.
	Some attempts have been made to create fact-checking pipelines with neural network based models.
	\citet{guo-etal-2022-survey} give an extensive overview of the tasks that need to be solved in the
	fact-checking pipeline and datasets used to
	train models for these tasks. An important part of fact-checking is claim detection,
	i.e., identifying content that should be checked because it contains some kind of claim. Rumor detection, i.e., the detection of unverified claims about any event, has been employed for this purpose, as, e.g., by \citet{guo2018rumor}. 
	Another important aspect of fact-checking is evidence
	retrieval. Recently, so called neural retrievers have emerged, which are
	models that, retrieve corresponding evidence given a query
	\citep{maillard2021multi}. Arguably, the most crucial part
	of fact-checking is verdict prediction, i.e., labeling the fact checked content. Simple models are binary classifiers which label a given claim as either true or false
	\citep{potthast2018stylometric}. However, a more fine-graded labeling is commonly used, e.g., \emph{false, barely-true, mostly true}, etc.~\citep{wang-2017-liar}. The final step in the fact-checking pipeline is to generate a justification for the given prediction, as explainability plays an important role when some fact is predicted as false. For this purpose models are trained to produce a textual justification~\citep{atanasova-etal-2020-generating-fact}. \citet{chen2024combating} examine the opportunities and challenges of LLMs in tackling the task of fact-checking. LLMs have the ability to generate fake news on a large scale, which, according to the latest findings, is very difficult to distinguish from real news \citep{chen2024can}. Nevertheless, the great advantage of using LLMs for fact-checking is that external knowledge can be fed into them. Automated web searches can also help ensure that the models have access to up-to-date information.

    Discourse that is automatically fact-checked is more \emph{rational}, as the accuracy of claims increases and participants might be motivated to verify their positions. Current systems are effective at identifying and verifying check-worthy claims, especially if they have access to up-to-date information. However, it should not be forgotten that LLMs are capable of generating and spreading misinformation that sounds valid. 
    
    %\paragraph{Automatic Web Searches}
    %\label{par:aws}
    %\citet{bjarnason2024using} suggest a framework for .. automated web searches to \todo{ausformulieren oder rauswerfen?}

    %Rephrasing user comments: suggest more polite formulation like in \citet{argyle2023leveraging} or more precise comments like suggested by \citet{klein2025can}.
    
% NEW \paragraph{Generate Text.} The significant improvement of large language models like ChatGPT \citet{} allows   generate text for participants~\citep{doi:10.1073/pnas.2311627120}.
	
	\subsection{Structuring the Discussion}
	\label{sec:struc}
	Another group of NLP tasks deals with using the information contained in discussions to provide a clearer, more structured overview. 
	
	\paragraph{Argument Mining.}
    \label{par:argu}
	The aim of argument mining is to automatically recognize and classify argumentative structures in text.  \citet{boltuzic-snajder-2014-back} initially proposed the
	task of argument recognition and come up with an approach for argument-based opinion mining. %zu alt?
	In a follow-up work, \citet{boltuzic-snajder-2015-identifying} extended the task of
	argument recognition by suggesting an unsupervised method to identify the most
	prominent argument in an online discussion. \citet{harly2022cnn} use a
	convolutional neural network (CNN)-BERT architecture to measure inter-argument
	agreement in online discussions. \citet{falk-lapesa-2022-reports} try to recognize 
	the utilization of personal experience in argumentation.
	One sub-task of argument mining is to select a set of keypoint arguments 
	from a collection of given arguments~\citep{friedman-etal-2021-overview, alshomary-etal-2021-key}.
	%\citet{alshomary-etal-2021-key} use machine learning methods, e.g. contrastive learning 
	%and extractive argument summarization to tackle the tasks of matching arguments to keypoints and
	%generating keypoints from given arguments. \emph{rationality},
	%\emph{interactivity}
	By extracting and displaying the key arguments of a discussion to participants, they can inform themselves about the discussion, keep an overview and form their own opinion.
	This contributes to a more informed, thus more \emph{rational} discussion.
	
	\paragraph{Duplicate Recognition.}
    \label{par:dup}
	Duplicate contributions, i.e., posts in discussions that 
	share the same idea, but use a different wording, 
	constitute a significant challenge for participants and initiators of 
	large-scale online participation processes.
	
	\citet{10.1145/1146598.1146663} tackle the task of recognition of near duplicate entries for e-rulemaking and propose an algorithm called DURIAN. There is not much recent work on identifying duplicate posts in online discussions and participation processes. However, the task of identifying semantically similar texts, which is also known as paraphrase identification in the NLP community, is useful for many areas, reaching from the identification of duplicate questions in online forums~\citep{9166343} to detection of plagiarism~\citep{10141785}.
	\citet{zhou2022paraphrase} give a recent overview of methods and datasets for paraphrase identification.
	
	Automatically identifying and marking similar posts in discussions improves clarity and counteracts the problem of information overload,
	allowing for a more \emph{rational} and structured discussion.
	
	\paragraph{Dialogue Summarization.}
    \label{par:summ}
    Automated summarization of written text, or
	more specifically, dialogue between multiple people, is another task that supports
	online discussions. It is closely related to argument mining, but deals with other challenges, such as capturing all expressed perspectives without omitting any points of view. 
	Summarization can provide key points for participants that
	join an ongoing discussion. For example, \citet{10.1145/3411763.3451815} found that
	presenting participants with a report on given arguments
	can improve users' ability to make sense of the discussion and their
	perception of its quality. \citet{arana-catania-etal-2021-evaluation} evaluate and compare different ML models to summarize deliberative processes for non-English
	languages.
    To solve this task LLMs can be trained and utilized.
	For this purpose, \citet{fabbri-etal-2021-convosumm} present a benchmark dataset
	for conversation summarization called ConvoSumm.\footnote{\url{https://github.com/Yale-LILY/ConvoSumm}} \citet{DBLP:journals/corr/abs-2306-11932} use an LLM to generate summaries for Polis \citep{small2021polis}. They propose a collaborative approach between LLMs and humans to evaluate automatically generated summaries because they pose the risk to leave out points of view, especially if they incorporate any kind of bias, which is very critical in the context of political and socially relevant discussions. When generating summaries, models could focus on areas of agreement or disagreement among participants. We refer to this task as \emph{consensus detection}. Pointing out areas where participants agree can strengthen the discussion. Highlighting areas of disagreement can lead to a focused search for solutions \citep{NEURIPS2022_f978c8f3,DBLP:journals/corr/abs-2306-11932}. 

    Another sub-task of text summarization, is the \emph{simplification} of the used language~\citep{al2021automated}, enabling more people to understand what the discussion is about.
    
	For a comprehensive overview on recent methods and challenges of dialogue summarization, 
	see the work of \citet{jia2022taxonomy}. Simple summaries of debates at the touch of a button could increase the \emph{equality} of online deliberation and motivate people to participate. 
    
	\paragraph{Structure Extraction.}
    \label{par:struc}
	Extracting the structure of online discussions is beneficial for facilitation
	tools in order to gain further insight of the ongoing discussion and extract
	important parts. The task involves identifying the relationship between messages and the flow of the conversation.
    There are different approaches to structure extraction. It can be solved with graph-based models, since conversations often exhibit tree or graph like structures in which posts are nodes and the edges represent their relationships. The process of structure extraction involves two main steps. First, nodes are classified, and then their relationships are predicted. \citet{suzuki2021structure} present
	a node classification approach using graph attention networks
	\citep{veličković2018graph}. %VISUALIZATION?.

    \citet{gupta2025civicparse} propose CivicParse, a benchmark and a pipeline for structured online deliberation. The pipeline is two-staged, first extracting distinct points of discussion that are classified with a content type, e.g., if the commenter proposes a solution to the discussion and a stance \emph{(pro, con)}. 
    \citet{10.1145/3672608.3707925} use an LLM to automatically embed a user's response in a deliberation map. The task for the LLM is to identify questions, proposed solutions and arguments, as well as the stance of a given argument to create deliberation maps that make it easier for participants to collaboratively discuss questions and find solutions for them. 

    Structure extraction is usually a combination of extraction and classification and contributes to a clearer, more \emph{rational} discussion. The models presented solve this task with high precision. 
	
	\paragraph{Topic Modeling.}
    \label{par:top}
	Topic modeling is the automated identification of the topic being discussed in a text.
	\citet{sun2023convntm} present a topic model using deep learning methods that is tailored for conversations.
	Automatically identifying the topic of a written post, e.g., on citizen platforms that gather ideas for improving cities, makes it easier for participants to 
	gain an overview and find posts they are interested in.
	Furthermore topic modeling could also help to find similar proposals to merge or suggest to other participants. In their experiments on Polis, \citet{DBLP:journals/corr/abs-2306-11932} show that prompting LLMs with topic modeling is an effective way to support and structure discussions. They use Anthropic's Claude \citep{DBLP:journals/corr/abs-2112-00861} to not only categorize the comments but also to produce two levels of topic labels. The authors emphasize that the task can be handled very well by LLMs and they consider the task less critical. Errors in topic assignments do not impact the discussion negatively, but a successful categorization can help to structure large-scale discussions.

    In addition to a better structure, assigned topics are also helpful to recommend contents to participants they might be interested in. This saves participants a lot of time and promotes exchange between people who have similar interests and pursue similar goals. This could significantly increase \emph{interactivity} and facilitate participation in general.
    
	\subsection{Sentiment Analysis}
	\label{sec:sent}
	Sentiment analysis, also referred to as opinion mining, is the automated
	identification of a users feelings, opinion and other subjective elements in
	text. While traditional sentiment analysis focuses on categorizing text as \emph{positive}, \emph{negative}, or \emph{neutral} and is often used to assess the overall tone of discussions as an evaluation step, sentiment signals in online discourse can also serve as a strong indicator of hateful or harmful content \citep{10.1007/978-3-031-43412-9_33}.

    \citet{su151410828} employ a Transformer-based architecture \citep{NIPS2017_3f5ee243} to address the task of politeness detection. A recent survey by \citet{sharma2025review} provides an extensive overview of sentiment analysis, including its core sub-tasks, benchmark datasets, and unresolved challenges.
    
	\paragraph{Hate Speech Detection.}
    \label{par:hate}
	Hate speech, toxic language and incivility often occur when people with
	opposing opinions discuss controversial topics. Civility is a central
	aspect of deliberation, as participants should feel heard and equally accepted 
	in a conversation~\citep{friess2015systematic}.
	Hate speech detection is the task of automatically identifying whether a given piece 
	of text contains hateful or offensive language. A widely used ML-based approach for recognizing toxic language in online discussions
	is Perspective.\footnote{\url{https://perspectiveapi.com/}} Perspective is a free to use
	application programming interface (API) that employs ML models to compute a
	toxicity score for user-generated comments.
	Hate speech detection encompasses several sub-tasks such as sarcasm~\citep{buaroiu2022automatic}, racism~\citep{9684891} and sexism~\citep{schutz2021automatic} detection.
	For an overview of recent methods and datasets for textual hate speech detection,
	we refer to the work of \citet{alkomah2022literature}.

    Effectively recognizing and moderating hateful comments has been shown to significantly improve the climate of discourse \citep{10.1145/3543507.3583275} and, consequently, increase overall \emph{civility}.
	
	\paragraph{Stance Detection.}
    \label{par:stance}
	Stance detection is a sub-task of sentiment analysis which aims to predict an author's
	position with regard to a given topic or central
	question (e.g. distinguishing between \emph{in favor} and \emph{against}). 
	It is, among other applications, commonly used to identify positions in political online discussions~\citep{WALKER2012719}. 
	Stance detection plays an important role for a variety of downstream tasks such as
	debate summarization and structuring. 
    A recent in-depth survey on the task of political viewpoint identification is given by \citet{doan2022survey}.
	%\citet{mascarell-etal-2021-stance} present a manually annotated dataset for
	%emotion recognition and stance detection in German news articles.\footnote{\url{https://github.com/mediatechnologycenter/CHeeSE}} 
    \citet{854df4346d18432ab8328fdeb08dc906} present the multilingual X-Stance dataset for stance detection in political contexts.\footnote{\url{https://github.com/ZurichNLP/xstance}}
    
	In the context of online discussions and online participation, the automatic detection of an authors
	position could be leveraged to structure the debate more effectively. For instance, it enables filtering and grouping of arguments to improve overview, or even to expose participants
	to opposing viewpoints and encourage exchange of opinions to potentially enhance \emph{interactivity}. 
	
	%\citet{math11092161} train transformer models on different stance detection
	%datasets of public discourse. The anticipated effect of stance detection
	%strongly depends on how the predictions are used.
	
	\subsection{Discourse Quality Measurement}
	\label{sec:qual}
    In contrast to filtering out rule violations in discussions, highlighting particularly high-quality contributions is another moderation strategy. Marking posts in discussion that fulfill deliberative
	standards, motivates participants to write better comments, improving the overall discourse quality~\citep{10.1145/3484245}.
    
	ML models that can assess the quality of a contribution
	could be used not only for this task, but also to evaluate tools and systems that aim to improve deliberation in
	online context. We provide an overview of previous approaches in the following.
    
	\paragraph{Measuring Deliberative Quality.}
    \label{par:qual}
	DelibAnalysis, developed by \citet{doi:10.1177/0165551519871828} is a framework to
	analyze discussion quality based on the discourse quality index (DQI)
	\cite{steenbergen2003measuring}.
	The DQI is a popular measure for deliberative quality. By now, other measures have been developed for different factors of quality, e.g., the listening quality index~\citep{scudder2022measuring} and the deliberative reasoning index~\citep{ercan2022research}. \citet{behrendt-etal-2024-aqua} propose a deliberative quality score for comments that is called AQuA, which is composed of 20 individual dimensions, such as justification and references from other users.
	
	\citet{fournier2021big} tackle the task of quality assessment with different ML
	techniques and predict a developed version of the DQI. They point out that for training these models, annotated datasets of high quality are needed. One dataset that includes contributions of a European-leveled poll annotated with the DQI, is EuroPolis~\citep{gerber2018deliberative}.
	In their work, \citet{falk-lapesa-2022-scaling} show how to augment data annotated for
	deliberative quality prediction in order to avoid class imbalances.
    
	\paragraph{Argument Quality Prediction.}
    \label{par:argqual}
	A subtask of measuring the overall deliberative quality of a discussion is to measure the quality of given arguments. Automatically accessing a score for
	given arguments can also be used to support human moderators in creating
	meaningful summaries. \citet{van-der-meer-etal-2022-will} try different
	approaches based on GPT-3~\citep{brown2020language} and
	RoBERTa~\citep{liu2019roberta} to predict the validity and novelty of arguments.

    Exchanging clear and convincing arguments is essential in deliberation, especially on socially relevant topics. Finding well-formulated and logically consistent arguments improves the overall \emph{rationality} of a discussion and helps with the decision-making process. LLMs could be used not only to measure and predict argument quality, but also improving the quality of written arguments \citep{wachsmuth-etal-2024-argument}. \citet{rescala-etal-2024-language} examine how well LLMs can recognize the quality of arguments and choose persuasive arguments based on demographics of the target group. They find that LLMs perform on a level similar to humans, which turned out to be not particularly high. Nevertheless, the authors see the risk that LLMs could spread misinformation based on particularly convincing arguments.

    For a recent overview on the topic, see the work of \citet{wachsmuth-etal-2024-argument}.
    
	\subsection{Evaluation of Participation Processes}
	\label{sec:evaluation}

	In the final phase of an online participation project, municipalities are faced with the
	challenge to evaluate the data they collected and transform it into political action. Evaluation is a time-consuming
	task in which NLP methods can make a noticeable difference~\citep{10.1145/3603254}. As previously mentioned, many tasks, such as sentiment analysis and text summarization, are used for evaluation, and will not be explained in detail again. In the following, we review previous work on evaluating participation processes using NLP methods.
	
	An important part of the evaluation is gaining insight into citizens' proposed arguments. \citet{liebeck2016airport} automatically extract components of arguments from a participation process regarding a German airport. \citet{romberg-conrad-2021-citizen}
	also follow this approach for a mobility-related urban planning in Germany. 
	They examine different ML models to recognize argumentative structures
	and classify them.
	\citet{9458376} use rule-based sentiment analysis with lexicons to calculate
	sentiment scores (either positive, neutral, or negative) for civic participation
	comments. Their work establishes the core functionality for the Civic CrowdAnalytics
	tool. Other tools for the automated evaluation of participation processes are described
	in Section \ref{subsec:eval}. For a recent overview of NLP methods for the evaluation of public participation projects, see the work of \citet{10.1145/3603254}.

    %\citet{borchers2024designing} propose 

    The AI-supported evaluation of online discussions and participation processes does not directly improve the discussions themselves. Nevertheless, if carried out precisely, taking into account all the opinions gathered and deriving actionable results from them, it is a very important step which can lead to greater trust in administrations. Well-evaluated processes increase citizens' motivation to participate in participatory processes. Ensuring that all voices are heard guarantees \emph{equality} of all participants in retrospect. As with every other aspect concerning the use of AI tools to support discussions, these tools should be adopted to support human actors, and not replace them. 

    \begin{table}
	\centering
    \small
	\begin{tabular}{lll}
		\hline
		Tool & Link & Task\\
		\hline
        All Our Ideas & \scriptsize{\url{https://all-our-ideas.citizens.is/}} & Engagement Tool\\
        bcause~\citep{10.1007/978-3-032-05008-3_19} & \scriptsize{\url{https://bcause.app}} & Discussion Platform\\
		Civic CrowdAnalytics \citep{10.1145/2994310.2994366} & \scriptsize{\url{https://github.com/ParticipaPY/civic-crowdanalytics}} & Discussion Evaluation\\
		CollAgree \citep{ito2014collagree} & - & Discussion Platform \\
		CommunityPulse \citep{10.1145/3461778.3462132} & \scriptsize{\url{https://communitypulse.cs.umass.edu/}} & Discussion Evaluation\\
		CONSUL \citep{10.1145/3452118} & \scriptsize{\url{https://consuldemocracy.org/}} & Citizen Participation Platform\\
        ConvoKit~\citep{chang-etal-2020-convokit} & \scriptsize{\url{https://github.com/CornellNLP/ConvoKit}} & Conversation Analysis Toolkit\\ 
        Coral & \scriptsize{\url{https://coralproject.net/}} & Commenting Tool\\
		Debate Hub & \scriptsize{\url{https://debatehub.net/}} & Discussion Platform\\
        DebateVis \citep{9331282} & \scriptsize{\url{https://debate-vis.xiaohk.vercel.app/}} & Visualization \\
		Decidim & \scriptsize{\url{https://decidim.org/}} & Citizen Participation Platform\\
        Deliberatorium & \scriptsize{\url{https://deliberatorium.org/}} & Discussion Platform\\
		%Deliveratorium & \\
		%D-Agree \citep{ito2022agent} & \small{\url{https://d-agree.com/site/en/}} &\\
        DIPAS \citep{lieven2017dipas} & \scriptsize{\url{https://dipas.org/en}} & Citizen Participation Platform\\
		Discourse & \scriptsize{\url{https://www.discourse.org/}} & Discussion Platform\\
        %DATS \citep{schneider-etal-2023-wise} & \small{\url{https://dats.ltdemos.informatik.uni-hamburg.de/}}& Discourse Analysis\\
        Go Vocal (former CitizenLab) & \scriptsize{\url{https://www.govocal.com/}} & Discussion Platform\\
		LiteMap & \scriptsize{\url{https://litemap.net/}} & Visualization \& Evaluation\\
		%LiquidFeedback & \url{https://liquidfeedback.com}\\
		Opinion Space \citep{10.1145/1753326.1753502} & - & Visualization\\ %\small{\url{http://opinion.berkeley.edu/}} &  \\
		%Packback & \url{https://www.packback.co/} \\
        Policy Synth \citep{bjarnason2024using} & \scriptsize{\url{https://policy-synth.ai/}} & Smarter Crowdsourcing\\
		Polis \citep{small2021polis} & \scriptsize{\url{https://pol.is/home}} & Discussion Platform\\
		Potluck \citep{lieu2022bring} (Prototype) &- & Discussion Platform\\
		%Regulation Room & \url{http://regulationroom.org/}\\
		RHETORiC \citep{10.1145/3491101.3503560} & \scriptsize{\url{https://www.rhetoric.tech/user/login}}& Commenting Tool\\
		%Smart Participation (former Regulation Room) & \url{http://smartparticipation.com/}\\
		%Stanford Online Deliberation Platform \citep{} & \small{\url{{https://stanforddeliberate.org/}} & Discussion Platform\\	% Video Platform
			%Sequent & \url{https://sequentech.io/}\\
		ThreadReconstructor \citep{el2018threadreconstructor}&- & Visualization\\
		%VisArgue \citep{ElAssady2016VisAr-37650} & \small{\url{https://visargue.lingvis.io/}} & Visualization\\
		Your Priorities & \scriptsize{\url{https://yrpri.org}} & Discussion Platform\\
		\hline
		\end{tabular}
		\caption{List of discussion platforms mentioned in this work.}
		\label{tab:tools}
	\end{table}

	\section{Overview of NLP in Practice}
	\label{sec:platforms}
	In the following Section we want to take a closer look on tools and platforms that integrate NLP or more general ML to enhance deliberative processes.
	
	\subsection{Discussion Platform Software}
	There is a wide range of different online discussion and participation platforms
	that include some kind of algorithmic facilitation, both open source and
	proprietary. In the following, we solely regard \emph{open source}, or at least \emph{free to use} non-profit platforms, as the transparency and accessibility of platforms is especially important for democratic processes.
	A list of all mentioned platforms with their link, if available, can be found in Table \ref{tab:tools}. For a list that also includes proprietary platforms, see the Democracy Technologies Database\footnote{\url{https://democracy-technologies.org/database/}}.\\

    \emph{\textbf{All Our Ideas}} is an engagement tool that is used to co-create a ranked list based on public input (a so called wiki survey). The tool enables ideas for a given question to be gathered and voted on. An integrated generative AI can automatically generate suitable graphics for the survey and also suggest answers that participants can vote on. \citet{gambrell2025ai} examines how an AI Task Force from the State of New Jersey was able to effectively use All Our Ideas to develop actionable policy recommendations. \\
	
	\emph{\textbf{bcause}} is a platform for accessible, structured and decentralized online
	deliberation, developed by \citet{10.1007/978-3-032-05008-3_19}.
	It includes automatic transcript analysis, argument mining and intelligent feedback aggregation and automated reporting with visualization tools. The authors aim to build a discussion environment with an integrated analytic engine that allows for more qualitative online debates.
	An experiment conducted by the research group showed that automated created reports can help users to make sense of an ongoing discussion and also improves the overall
	debate quality~\citep{10.1145/3411763.3451815}. Recently, the platform has been expanded by an argument mining tool \citep{elguendouze-etal-2025-am4dsp}.\\
	
	\emph{\textbf{COLLAGREE}} (short for collective or collaborative
	agreement) is an online discussion platform that supports consensus decision-making, developed by \citet{ito2014collagree,ito2015incentive}. COLLAGREE implements
	functions to support discussion facilitators. These functions
	include a text-based sentiment analysis, an automated
	agreement/ disagreement prediction for each post, and an automatic generation of highlighted keywords. \citet{ito2014collagree} 
	also added an incentive mechanism that rewards participants for the effectiveness of their posts.\\
	
	\emph{\textbf{CONSUL}} is another citizen participation tool for debates, proposals, budgeting and polls. \citet{10.1145/3452118} develop and evaluate
	different NLP tools for the platform 
	to counteract the problem of information overload that is often
	observed in digital mass participation. In their project, the authors mainly
	focus on tackling information overload with regard to proposals that can be added
	to the platform by citizens. The modules integrated in Consul categorize proposals, suggest relevant proposals for users so that they can support them, group citizens by their interests to
	encourage the exchange among citizens and summarize the comments that were
	added to one proposal. \citet{davies2021evaluating} use the developed
	NLP methods in Consul to facilitate participatory budgeting processes in
	Scotland. In the future, the platform will be expanded to include an open-source LLM-based assistant.\\

    \emph{\textbf{Coral}} is a tool for large communities, primarily built for journalism. The tool is supported by many features that simplify effective moderation, such as highlighting and filtering comments and give feedback to participants. The developers at Coral rely on algorithmic support for moderation, but firmly believe that human moderators should always make the final decision.\\
    
    \emph{\textbf{Debate Hub}}~\citep{quinto2021designing} is a tool to share ideas and discuss pro and contra arguments. It has been developed by the Open University's Knowledge Management Institute. The tool, i.a., includes a grouping mechanism to build discussion groups, a moderator toolbar and a visualization dashboard to better structure and understand the ongoing debate.\\

    \emph{\textbf{Decidim}} is a large-scale digital platform for citizen participation which is continuously being developed by a whole community. The project was initiated by the Decidim Free Software Association. The current version includes services for spam detection to support moderators on the platform and the possibility to incorporate machine translations.\\

    \emph{\textbf{Deliberatorium}} \citep{klein2011harvest} is a discussion platform, based on large-scale argumentation and
	argument mapping. Participants can either post an issue, an idea or a pro
	or contra argument for an idea or another argument. The discussion is structured
	in a tree-like network. Moderators verify each post and check if they
	follow the guidelines to ensure that every gathered idea is unique and people
	can support each others ideas. Currently, LLM support for the platform, including argument mining, answer clustering and comment rephrasing suggestions, is being evaluated, and an extension of the Deliberatorium is being developed \citep{klein2025can, 10.1145/3672608.3707925}.\\

    \emph{\textbf{DIPAS}}~\citep{lieven2017dipas} is a digital citizen participation tool, developed by Hamburg's Department of Urban Development and Housing in Hamburg, the State Office for Geoinformation and Surveying and the HafenCity University's CityScienceLab. The tool has two components, an online module and an on-site module. The online component is a participation platform where citizens can submit proposals, questions and criticism, which can be linked to locations in the city via geodata. The on-site component is used for face-to-face meetings where data is gathered on digital touch tables. DIPAS Analytics is an NLP toolbox that supports the evaluation and analysis of text contributions in DIPAS. The toolbox includes methods to extract key points and categorize them and links text and geodata.\\
	
	\emph{\textbf{Discourse}} is a community platform including features for civilized online discussions. The platform has several AI modules build in, e.g., automatic topic recommendation based on semantic similarity, an assistant to write a new topic that proofreads the users' text, suggests titles and translates it to English, a toxicity detection module and a summarization tool that summarizes discussed topics. More features supported through AI are currently planned.\\

    \emph{\textbf{Go Vocal}} (former CitizenLab) is a citizen participation platform, developed by an eponymous company, founded in Brussels. The software is used worldwide for citizen participation processes and polls. The platform includes an algorithm that is able to automatically analyze proposals of citizens to extract keywords and tag them. Additionally, NLP methods are used to identify posts with similar topics and summarize them. This helps participants to find topics of interest and decrease information overload on the platform.\\
	
    \emph{\textbf{Policy Synth}}~\citep{bjarnason2024using} integrates the All Our Ideas and Your Priorities platforms and LLMs, such as GPT-4~\citep{achiam2023gpt} for so called Smarter Crowdsourcing. The tool is able to automatically formulate, rank and prioritize both socially relevant issues and potential solutions. It helps to combine recommendations made by experts and solutions proposed by LLMs based on web searches to enhance decision-making and idea gathering processes. A case-study by \citet{gambrell2025ai} demonstrates how these tools could be used effectively to support policy-makers come up with recommendations for socially relevant issues.\\
	
	\emph{\textbf{Polis}}~\citep{small2021polis} is a system for
	idea gathering and analysis supported by ML, developed by the Computational
	Democracy Project.\footnote{\url{https://compdemocracy.org/}} Participants can add comments to specific topics, which are randomly suggested to other users who can vote on them.
	This creates a sparse comment matrix. After the conversation, the data matrix
	is ready to be exported and analyzed. Polis solely relies on language
	independent methods calculated on the data matrix like PCA
	\cite{pearson1901liii}, UMAP~\citep{2018arXivUMAP} and different clustering
	methods. Recently, the potential and risks of integrating LLMs into Polis have been examined by \citet{DBLP:journals/corr/abs-2306-11932}. They use Anthropic's Claude \citep{DBLP:journals/corr/abs-2112-00861} for tasks such as topic modeling, summarization, and generating textual cues to guide conversations for human facilitators. They conclude that LLMs have the potential to improve Polis discussions effectively. However, they ultimately emphasize the importance of human involvement in human-AI collaboration. \\

	\emph{\textbf{Potluck}}~\citep{lieu2022bring}, a commenting system for large scale online discussions, aims to allow for more constructive and inclusive online discussions. It is named after a form of dinner where everyone brings and shares their food in a gathering. The systems includes a module for automatic summarization using a pre-trained DistillBART model.\footnote{\url{https://huggingface.co/sshleifer/distilbart-cnn-12-6}} For aggregation of similar posts they use SentenceBERT~\citep{reimers-gurevych-2019-sentence} and the Perspective API~\citep{perspectiveapi} to screen inputs for toxicity. Until now, there is only a prototype of the tool available.\\
	
	\emph{\textbf{RHETORiC}} (an acronym for Reducing Hate through Editorial Tools for Online Reactions and
	Comments) by \citet{10.1145/3491101.3503560} is a
	discussion tool that supports civil participation in news comment sections. The main features of RHETORiC are: real-time feedback on the quality of
	the written comments, proposal of similar arguments, the possibility to like
	other comments and a clustered view of all comments based on their similarity.
	Comments are also automatically classified and sent to pre-moderation, if they
	appear to be uncivil. Experiments showed that the made design choices had an
	impact on the comment quality and the overall engagement~\citep{10.1145/3491101.3503560}. However, participants
	did not always agree with the decisions of the ML algorithm. Thus
	the authors plead to further develop and utilize explainable ML algorithms that
	are more transparent in their decisions.\\
	
	\emph{\textbf{Your Priorities}} is an online platform for idea generation, deliberation and
	decision making, developed by the Citizens Foundation.\footnote{\url{https://citizens.is/}} The platform includes a
	detector for inappropriate content, a module for automatic translations and a
	recommendation algorithm that suggests posts that most likely are interesting to a user. It also implements a fraud detection system and an analysis tool to cluster ideas.\\

    There is another branch of research that develops tools especially for discussion analysis and annotation, see e.g. the work by \citet{fischer-etal-2024-extending}, which we did not include in our work.
	
	% \citet{begen2020} develops a classifier for the participation platform "Our
	% St. Petersburg" to
	
	% Other platforms that are open source and free to use for public participation,
	% but do not include supporting algorithms are: adhocracy+
    
	\subsection{Visualization Software}
	Another approach to gather insights on political debates is to visualize them, 
	in order to navigate through conversations and explore different aspects
	visually. In the following we describe tools that analyze conversation through visualization.\\% \citet{naerland2023towards}

    \emph{\textbf{DebateVis}}~\citep{9331282} is a discussion visualization tool that helps non-expert users to navigate and analyze transcripts of political debates. It can, e.g., be used in preparation for elections to inform voting decisions. Given the transcript for a debate, DebateVis provides the user with an interaction graph, marking the mentions of other candidates and topics in the discussion. \\
    
    \emph{\textbf{LiteMap}} is a tool that collects, visualizes, analyzes, and summarizes the content of online discussions across different platforms~\citep{10.1145/3461564.3461584}. LiteMap helps to gather ideas and pro and contra arguments from public debates. The tool incorporates a booklet function to save web content and a website to create and share argument maps.\\
	
	\emph{\textbf{Opinion Space}}~\citep{10.1145/1753326.1753502} is another tool that enables users to collect and visualize opinions online. Comments are depicted as dots on an interactive map, with nearby dots representing similar opinions and distant dots representing dissimilar opinions. The Opinion Space interface should counteract the problems of typical list-based discussion forums, in which users tend to lose track of the discussion.\\
	
	\emph{\textbf{ThreadReconstructor}}, presented by \citet{el2018threadreconstructor}, is a tool to reconstruct threads and untangle conversations in large online discussions.
	The tool is specialized on transcripts of discussions in online forums, where no threaded reply structure is given.\\
	
	%%\paragraph{StanceVis Prime} StanceVis Prime is a visualization software to analyze 
	%%and visualize sentiment and stance of documents. The platform is not limited to discussion data, it can be
	
%	\emph{\textbf{VisArgue}}~\citep{ElAssady2016VisAr-37650} is a visualization toolbox to further analyze and visualize political debates,
%	by \citet{ElAssady2016VisAr-37650}.
%	It combines different tools, e.g., the Lexical Episode Plots~\citep{10.1093/llc/fqv033} to cluster and
%	visualize the most significant terms of a conversation and ConToVi~\citep{} for an
%	analysis of the specific patterns of participants.
	
	\subsection{Evaluation Software}
	\label{subsec:eval}
	Besides ML supported platforms, there are also tools to evaluate and analyze
	citizen participation processes which are presented in the following. These tools are particularly useful for initiators of participation processes.\\
	
	\emph{\textbf{Civic CrowdAnalytics}} is a web application,
	developed by \citet{10.1145/2994310.2994366} to analyze data gathered from
	crowdsourcing processes in the context of policy making. The tool includes topic
	categorization via concept extraction, sentiment analysis~\citep{9458376}
	(indicating positive, neutral or negative sentiment) and shows which words occur
	often in the analyzed text.\\
	
	\emph{\textbf{CommunityPulse}} is an ongoing research project to
	explore comments from community participation processes and visualize them
	\citep{10.1145/3461778.3462132}. The conduction of expert interviews revealed
	that for organizers of civic participation it is most important to keep an overview of the gathered data, both on a high level as well as on the individual comment level. Furthermore it is important to grasp the overall sentiment and to understand the main topics that were discussed. 
	Based on these findings, CommunityPulse offers three different views. One for visualizing the discussion, the topics and the sentiment of the participants on a high level. A view to analyze comments in more detail and a view to directly compare contributions.\\

    \emph{\textbf{ConvoKit}}~\citep{chang-etal-2020-convokit},  is a toolkit that combines several methods for text analysis. With the help of ConvoKit, linguistic features and social phenomena can be extracted from conversations, such as politeness, or rather impoliteness of the used language, linguistic diversity and redirection of conversation flow. A model that is trained to forecast the outcomes of conversations is also included.\\

    Table \ref{tab:tools} lists all mentioned tools and a link to their website or GitHub Repository, if available.

    \begin{table}[h!]
		\centering
        \small
		\begin{tabularx}{\textwidth}{lX}
			\hline
			Machine Learning Method & Reference \\
			\hline
            Adapter~\citep{pfeiffer-etal-2021-adapterfusion} & \citet{falk-lapesa-2023-bridging, behrendt-etal-2024-aqua}\\
			BART~\citep{lewis-etal-2020-bart} & \citet{arana-catania-etal-2021-evaluation}\\
			BERT~\citep{devlin_etal_2019_bert} & \citet{arana-catania-etal-2021-evaluation, anastasiou-2022-enrich4all, 10.1145/3461778.3462132, falk-etal-2021-predicting, electronics13030544, katsaros2022reconsidering, electronics13030544, falk-lapesa-2022-reports,fournier2021big}\\
			BiLSTM~\citep{10.1162/COLI_a_00295} & \citet{ito2022agent, guo2018rumor}\\
            BoW& \citet{sun2023convntm,fournier2021big} \\
			BTM~\citep{10.1145/2488388.2488514} & \citet{el2018threadreconstructor}\\
			%Case-Based Reasoning & \\
            Claude~\citep{DBLP:journals/corr/abs-2112-00861} & \citet{DBLP:journals/corr/abs-2306-11932}\\
            %CNN & \citet{9166343} \\
            CNN-BERT~\citep{harly2022cnn} & \citet{harly2022cnn}\\
			Decision Trees & \citet{10.1145/2994310.2994366}\\
            DeepSBD~\citep{9505695} & \citet{9505695}\\
			DistillBART & \citet{lieu2022bring} \\
            DistillBERT \citep{sanh2020distilbertdistilledversionbert} & \citet{atanasova-etal-2020-generating-fact}\\
            ECGA \citep{8789863} & \citet{romberg-conrad-2021-citizen}\\
            fastText \citep{joulin-etal-2017-bag} & \citet{romberg-conrad-2021-citizen}\\
            GCR-NN~\citep{9684891} & \citet{9684891}\\
			GloVe Embeddings~\citep{pennington2014glove} & \citet{10.1145/3452118, 9166343}\\
			GPT-3~\citep{brown2020language} & \citet{van-der-meer-etal-2022-will, doi:10.1073/pnas.2311627120, 10.1145/3613905.3650828,rescala-etal-2024-language}\\
            GPT-4~\citep{achiam2023gpt} & \citet{bjarnason2024using, 10.1145/3672608.3707925,rescala-etal-2024-language} \\
			Graph Attention Networks~\citep{veličković2018graph} & \citet{suzuki2021structure}\\
			IHTM~%Incremental Hierarchical Topic Modeling 
			\citep{assady2015incremental} &\citet{el2018threadreconstructor}\\
			K-Means & \citet{10.1145/2994310.2994366}\\
            K-nearest-neighbor classification & \citet{liebeck2016airport}\\
			LDA~%Latent Dirichlet Allocation 
			& \citet{10.1145/3452118, 10.1145/3461778.3462132, el2018threadreconstructor, 9331282} \\
            %LSTM~\citep{10.1162/neco.1997.9.8.1735} & \citet{} \\
            Llama 2\citep{touvron2023llama2openfoundation} & \citet{rescala-etal-2024-language} \\
            Mistral 7B\citep{jiang2023mistral7b} & \citet{rescala-etal-2024-language}\\
			Naive Bayes & \citet{10.1145/2994310.2994366}\\
            No Language Left Behind~\citep{nllbteam2022languageleftbehindscaling} & \citet{electronics13030544}\\
			Non-Negative Matrix Factorization & \citet{10.1145/3452118}\\
			PCA~%Principal Component Analysis 
			\cite{pearson1901liii} & \citet{small2021polis, 10.1145/1753326.1753502}\\
			Perspective API~\citep{perspectiveapi}& \citet{10.1145/3543507.3583275}\\
            PMI-IR~\citep{10.3115/1073083.1073153} & \citet{ito2014collagree}\\
			Random Forest Classifier & \citet{doi:10.1177/0165551519871828, 10.1145/2994310.2994366, falk-etal-2021-predicting, liebeck2016airport}\\
			RoBERTa~\citep{liu2019roberta} & \citet{falk-etal-2021-predicting, van-der-meer-etal-2022-will}\\
			Sentence-BERT~\citep{reimers-gurevych-2019-sentence} & \citet{lieu2022bring}\\
			SWB~%Special Words With Background 
			\citep{NIPS2006_ec47a5de} & \citet{el2018threadreconstructor}\\
			SVM %Support Vector Machine 
			& \citet{10.1145/2994310.2994366, liebeck2016airport, romberg-conrad-2021-citizen}\\
			TF-IDF & \citet{10.1145/3452118, 10.1145/2994310.2994366} \\
			TextRank~\citep{mihalcea-tarau-2004-textrank} & \citet{10.1145/3452118}\\
            Transformer~\citep{NIPS2017_3f5ee243}&\citet{chang-etal-2020-convokit, sun2023convntm, su151410828, schutz2021automatic}\\
			UMAP~\cite{2018arXivUMAP} & \citet{small2021polis}\\
			VADER~\citep{hutto2014vader} & \citet{10.1145/2994310.2994366}\\
			\hline
		\end{tabularx}
		\caption{List of machine learning methods applied in online discussions and participation processes.}
		\label{tab:mltools}
	\end{table}

	\subsection{Used Machine Learning Methods}
	We gave an extensive overview of ML and NLP methods in existing tools for online debating and participation. Table \ref{tab:mltools} lists all ML methods and models
	that have been used throughout literature. While some works have utilized, mostly Transformer based~\citep{NIPS2017_3f5ee243} deep learning models such as BERT~\citep{devlin_etal_2019_bert} or RoBERTa~\citep{liu2019roberta}, most of the listed methods are classical ML methods such as support vector machines (SVMs) and K-Means. 
	
	\subsection{Datasets}
	
	High-quality datasets annotated by experts form the foundation of ML
	models that solve the tasks discussed in the given work.
	The lack of high-quality datasets remains a key challenge for training and utilizing models to facilitate processes in the public sector.
	However, we found datasets during our literature review that can be used for this purpose. The list of datasets and their respective tasks is provided in Table \ref{tab:datasets}. The results are limited to English-language datasets, or multilingual datasets that include English. Please note that this list is not exhaustive, and there are a lot of data sets for each task area. Here, we have limited ourselves to data sets in which conversations from online discussions or participation processes have been annotated. 

	\begin{table}[!ht]
		\centering
        \small
		\begin{tabularx}{\textwidth}{XX}
			\hline
			Dataset & Task \\%& Source\\
			\hline
            args.me \citep{ajjour:2019a}& Argument Search  \\
            ArguAna Counterargs \citep{wachsmuth:2018a} & Counterargument Retrieval\\
			Change my View~\citep{10.1145/2872427.2883081} & Persuasive Arguments\\% & Reddit\\
			CivicParse \citep{gupta2025civicparse} & Structure Extraction\\% & Deliberatorium\\
			CoFE \citep{barriere-etal-2022-cofe} & Multilingual Multi-target Stance Classification\\ %& CoFE\\
            Conversations Gone Awry \citep{chang-danescu-niculescu-mizil-2019-trouble} & Conversation Derailment Forecasting \\%& Reddit \& Wikipedia \\
            Cornell eRule-making Corpus \citep{park2012facilitative} & Argument Mining \\%& Regulation Room\\
            DEBAGREEMENT \citep{pougue2021debagreement}& (Dis)agreement Detection \\%& Reddit\\
			Debating Europe \citep{barriere2022debating}& Multilingual Multi-target Stance Classification \\%& Debating Europe\\
            Deliberation Bank \citep{zhu2025can}& Opinion Summarization \\
            DeliData \citep{Karadzhov2021DeliData}& Multi-Party Problem Solving \\
			Europolis Dataset \citep{gerber2018deliberative} & Deliberative Quality \\
			Internet Argument Corpus \citep{walker-etal-2012-corpus} & Several Dialogue and Argument tasks \\
            Factify \citep{mishra2022factify} & Fact-Verification \\
            Factify 2 \citep{suryavardan2022factify} & Fake News \& Satire News \\
            Jigsaw Toxic Comment Dataset \citep{jigsaw-multilingual-toxic-comment-classification} & Multilingual Toxic Comment Classification \\
            PerspectiveMod \citep{vecchi-etal-2025-perspectivemod} & Moderator Intervention Prediction \\%& Reddit \& Regulation Room\\
			StoryARG \citep{falk-lapesa-2023-storyarg} & Personal Stories in Arguments \\%& New York Times, Reddit \& Regulation Room \\
            %UKP \\
            Twitter Spam Detection Dataset~\citep{utkmls-twitter-spam-detection-competition} & 
            Social Spam Detection \\%& Twitter\\
            UMOD Dataset \citep{falk-etal-2024-moderation}& User Moderation \\
            Webis-TLDR-17 Corpus \citep{voelske:2017} & Text Summarization\\
            X-Stance \citep{854df4346d18432ab8328fdeb08dc906} & Stance Detection \\
			\hline
		\end{tabularx}
		\caption{List of annotated political online discussion and participation process datasets.}
		\label{tab:datasets}
	\end{table}

	\section{Key Challenges \& Open Problems}
    \label{sec:challenges}
	So far we mainly discussed opportunities for applications of ML in online discussions
	and participation processes. Still there are some key challenges and risks that need to be addressed.
	
	\paragraph{Lacking Accuracy.}
	The number of tasks that ML systems can solve in order to
	support municipalities and policy makers is growing. However, methods still need improvement to generate meaningful insights, e.g., in participation
	process data for evaluation purposes~\citep{10.1145/2994310.2994366}. For this purpose, annotated datasets of high quality are necessary. The latest research shows that LLMs trained on large amounts of data can perform these tasks very well in some cases \citep{DBLP:journals/corr/abs-2306-11932}. However, using LLMs also brings other risks, which we will discuss below.
	
	\paragraph{Induced Biases.} Automated processes that are used in social
	contexts always pose risks. This is especially true when tracing the
	transparency, explainability, plausibility, reliability or legitimacy of the
	applications is not possible. \citet{strauss2021deep} points out that it is necessary
	to keep potential biases in mind when developing methods and also when using
	them in practice. Precisely in the field of politics and online discussions
	where data to train models is taken from real conversations, it is likely that
	the data and subsequently also the models that learn from the data contain
	political biases, leading to unethical, unfair or even discriminating decisions~\citep{10.1145/3457607}.
	
	\paragraph{Manipulation.} The use of ML methods in environments with social
	impact such as political participation always has to be well considered. It
	has been shown that algorithms can lead to increased polarization,
	selective exposure~\citep{dylko2017dark} and influenced views. ML can
	help us make sense of big amounts of data and can be a supportive tool,
	but it should also be used in a transparent and ethical way. 
	
	\paragraph{Transparency.} We have seen a variety of tasks that can be 
	solved or at least supported through ML methods. Nevertheless, if it is not 
	clear for the participants how methods are used, what they do in the process
	and how they make decisions, it can have an impact on the process and 
	lead to trust issues~\citep{ORBi-ca9c331b-d6ec-426d-9db3-4a17bb7ab193, gambrell2025ai}. Explainable ML is an own branch of research that is
	studied by many researchers, yet it is not common to use ML methods that are
	transparent to those who use it. \
	
	\paragraph{Lack of Data.} In order to train models that support
	participation processes, up-to-date data on current issues and of high quality is necessary.
	Communes would profit from sharing their newest data, but data protection
	policies do not always allow free sharing of discussion data.

    \paragraph{Hallucination.} With the emergence of LLMs and their potential application in online discussions and participation processes, there is a high risk of models to hallucinate facts~\citep{10.1145/3703155}, or invent details when summarizing user opinions~\citep{DBLP:journals/corr/abs-2306-11932}. Incorrect facts and errors in summaries are very difficult to identify and can potentially distort a discussion.
	
	\paragraph{Evaluation.} There are many approaches from the field of
	ML to solve potential issues in online discussions and online
	participation processes. However, not many have been used and examined in
	practice, yet. It is still an open question whether the application of NLP methods can demonstrably contribute
	to improve online discussions and if they are perceived as helpful by
	participants. Controlled experiments could help to systematically evaluate the
	use of ML in online discussions and provide a better understanding on how these
	tools could be used in a supporting way~\citep{HELBING2023102061}.

    Many of these risks can be mitigated by keeping humans in the loop and leaving the opportunity to override and correct decisions made by ML models.
	
	\section{Conclusion}
	\label{sec:conlcusion}
	Civic participation is an important instrument to democracy to collectively solve problems of public concern, both on a local as well as on a global level. Today, more and more communities and municipalities make use of digital tools to include citizens in decision making. 
	While the digital space holds many opportunities, as it is neither location nor time-bound and enables humans from all over the world to contribute to important decisions, it also implies issues. It is important to
	never lose track of potential risks. Unfortunately, the use of algorithms is always twofold, as it holds
	potential for manipulation and abuse.
	
	%	The success of online discussions and online participation processes crucially 
	%	depends on the design of the chosen software tools.
	%	In this work we have summarized many possibilities where machine learning,
	%	especially NLP, can support processes and enhance deliberation.
	%	
	
	%In this work we discussed some of the key issue for large-scale online participation processes. 
	%We gave an overview of NLP tasks that have potential to be used and that
	%are already in use to support participants and initiators of online discussions 
	%while enhancing deliberative communication.
	
	%There is currently increasing research towards the use of AI in an educational context
	%\citep{10.1186/s41239-019-0171-0}. But there is not much work done for
	%discussion platforms or participation platforms. 
	
	With our work we hope to showcase the importance of deliberative elements in 
	online discussions, especially in political contexts. If more platforms
	implement designs that enhance respectful, inclusive conversations online,
	citizen councils and other democratic instruments hold great potential
	to come to thoughtful decisions and collectively fight crises that affect all
	people. A platform where design choices are gathered and scientifically evaluated
	is the Prosocial Design Network\footnote{\url{https://www.prosocialdesign.org/}},
	which is continuously developed further.
	
	ML has the potential to noticeably
	facilitate collective decision making, but it should always be seen as a
	supporting tool and not replace the work of humans. Nevertheless, it will be part of our lives as it is
	already today. Researchers should further investigate the use of
	ML, but also study side effects and risks that accompany it.

\bibliographystyle{ACM-Reference-Format}
\bibliography{sample-base}

%\appendix
%\newpage
%	\section{Appendix}
%\label{sec:appendix}
%\setlength\intextsep{-5pt}
%\subsection{Tools for Online Deliberation}
%In Section \ref{sec:platforms} we gave on overview on online discussion and
%participation tools that utilize some kind of machine learning support. 

\end{document}